\documentclass[dvipsnames]{article}

\usepackage[final]{openbmb_2025}

\usepackage{amsmath}
\usepackage{amssymb}
\usepackage{mathtools}
\usepackage{amsthm}
\usepackage{multirow}
\usepackage{makecell}
\usepackage{caption}
\usepackage{subcaption}
\usepackage{wrapfig}
\usepackage{graphicx}
\usepackage{pifont}
\usepackage{siunitx}

\usepackage[table]{xcolor}
\usepackage{tabularx}
\usepackage{array}
\usepackage{siunitx}
\usepackage{fvextra}
\usepackage{stmaryrd}
\usepackage{wrapfig} 

\usepackage[most]{tcolorbox}
\newtcolorbox[auto counter, number within=section]{examplebox}[2][]{%
    enhanced,
    breakable,
    fonttitle=\bfseries,
    title=#2,
    label={#1},
    colback=white,
    colframe=black!75,          
    colbacktitle=black!65,      
    coltitle=white
}

\newcommand{\code}[1]{{\color{leanKeyword}\texttt{#1}}}

\definecolor{leanKeyword}{HTML}{1F4E79}   
\definecolor{leanCommand}{HTML}{7A3E9D}   
\definecolor{leanTactic}{HTML}{0F766E}    
\definecolor{leanComment}{HTML}{6A737D}   
\definecolor{leanString}{HTML}{A05A2C}    
\definecolor{leanSymbol}{HTML}{374151}    
\lstdefinelanguage{lean4}{
    sensitive=true,
    alsoletter={_'},
    morecomment=[l]{--},
    morecomment=[s]{/-}{-/},
    morestring=[b]",
    morekeywords=[1]{
        import,open,namespace,section,end,variable,variables,universe,universes,
        theorem,lemma,example,def,abbrev,axiom,opaque,inductive,coinductive,
        structure,class,instance,where,deriving,syntax,macro,notation,
        scoped,local,mutual,partial,noncomputable
    },
    morekeywords=[2]{
        by,fun,let,have,show,suffices,from,if,then,else,match,with,
        do,for,in,return,try,catch
    },
    morekeywords=[3]{
        intro,intros,apply,exact,refine,rw,simp,dsimp,constructor,
        cases,induction,subst,rfl,assumption,ring,omega,linarith,
        norm_num,aesop,unfold,decide,contradiction,trivial,left,right,
        use,obtain,rcases
    },
    keywordstyle=[1]\color{leanCommand}\bfseries,
    keywordstyle=[2]\color{leanKeyword}\bfseries,
    keywordstyle=[3]\color{leanTactic}\bfseries,
    commentstyle=\color{leanComment}\itshape,
    stringstyle=\color{leanString},
    basicstyle=\ttfamily\small\color{leanSymbol},
    breaklines=true,
    breakatwhitespace=false,
    showstringspaces=false,
    columns=flexible,
    keepspaces=true,
    extendedchars=true,
    literate=
        {ℝ}{{$\mathbb{R}$}}1
        {ℂ}{{$\mathbb{C}$}}1
        {ℕ}{{$\mathbb{N}$}}1
        {ℤ}{{$\mathbb{Z}$}}1
        {ℚ}{{$\mathbb{Q}$}}1    
        {π}{{$\pi$}}1
        {��}{{$\mathcal{N}$}}1
        {ᶜ}{{$^{\mathrm{c}}$}}1
        {₀}{{$_0$}}1
        {₁}{{$_1$}}1
        {₂}{{$_2$}}1
        {₃}{{$_3$}}1
        {₄}{{$_4$}}1
        {₅}{{$_5$}}1
        {₆}{{$_6$}}1
        {₇}{{$_7$}}1
        {₈}{{$_8$}}1
        {₉}{{$_9$}}1
        {∈}{{$\in$}}1
        {∉}{{$\notin$}}1
        {⊆}{{$\subseteq$}}1
        {⊂}{{$\subset$}}1
        {∃}{{$\exists$}}1
        {∀}{{$\forall$}}1
        {≤}{{$\le$}}1
        {≥}{{$\ge$}}1
        {≠}{{$\ne$}}1
        {∧}{{$\wedge$}}1
        {∨}{{$\vee$}}1
        {¬}{{$\neg$}}1
        {→}{{$\to$}}1
        {↔}{{$\leftrightarrow$}}1
        {λ}{{$\lambda$}}1
        {·}{{$\cdot$}}1
        {⟨}{{$\langle$}}1
        {⟩}{{$\rangle$}}1
        {‖}{{$\Vert$}}1
        {ν}{{$\nu$}}1
        {α}{{$\alpha$}}1
        {β}{{$\beta$}}1
        {γ}{{$\gamma$}}1
        {δ}{{$\delta$}}1
        {ε}{{$\varepsilon$}}1
        {θ}{{$\theta$}}1
        {λ}{{$\lambda$}}1
        {μ}{{$\mu$}}1
        {ν}{{$\nu$}}1
        {π}{{$\pi$}}1
        {ρ}{{$\rho$}}1
        {σ}{{$\sigma$}}1
        {τ}{{$\tau$}}1
        {φ}{{$\varphi$}}1
        {ψ}{{$\psi$}}1
        {ω}{{$\omega$}}1
        {×}{{$\times$}}1
        {⊤}{{$\top$}}1
        {⊥}{{$\bot$}}1
        {•}{{$\cdot$}}1
        {∑}{{$\textstyle\sum$}}1
        {∏}{{$\prod$}}1
        {∫}{{$\int$}}1
        {ˢ}{{$^{\mathrm{s}}$}}1
        {∂}{{$\partial$}}1
        {⨅}{{$\bigsqcap$}}1
        {⨆}{{$\bigsqcup$}}1
}

\usepackage{newunicodechar}

\usepackage[utf8]{inputenc} 
\usepackage[T1]{fontenc}    
\usepackage{url}            
\usepackage{booktabs}       
\usepackage{amsfonts}       
\usepackage{nicefrac}       
\usepackage{microtype}      

\usepackage{color, soul}
\usepackage[most,breakable]{tcolorbox}
\usepackage{xcolor}         
\usepackage{listings}
\usepackage{amssymb} 

\usepackage[labelfont=bf]{caption}
\usepackage{enumitem}
\usepackage{tablefootnote}
\usepackage{threeparttable}
\usepackage{tabularx}
\usepackage{booktabs}
\usepackage{colortbl} 
\definecolor{lightgray}{gray}{0.95}

\usepackage{fancyhdr}
\usepackage{wrapfig}
\renewcommand{\headrulewidth}{1pt}
\makeatletter
\def\headrule{{\if@fancyplain\let\headrulewidth\plainheadrulewidth\fi
\hrule\@height\headrulewidth\@width\textwidth \vskip-\headrulewidth}}
\makeatother

\definecolor{BMBDarkBlue}{HTML}{315EFE}
\definecolor{BMBLightBlue}{HTML}{00D3ED}

\usepackage[colorlinks, linkcolor=RoyalBlue, anchorcolor=RoyalBlue, citecolor=RoyalBlue, urlcolor=RoyalBlue]{hyperref}
\usepackage{amsthm}

\newtcolorbox{mytheorem}{
  colback=gray!5, 
  colframe=gray!80, 
  boxrule=0.5pt, 
  arc=4pt, 
  left=4pt, 
  right=4pt, 
  top=4pt, 
  bottom=4pt, 
}

\title{MA-ProofBench: A Two-Tiered Evaluation of LLMs for Theorem Proving in Mathematical Analysis}

\author{%
Lushi Pu\textsuperscript{\rm 1}, 
Weiming Zhang\textsuperscript{\rm 2},
Xinheng Xie\textsuperscript{\rm 1},
Zixuan Fu\textsuperscript{\rm 2},
Bingxiang He\textsuperscript{\rm 2},\\
\textbf{
Hongya Lyu\textsuperscript{\rm 1}, 
Xin Li\textsuperscript{\rm 1},
Jie Zhou\textsuperscript{\rm 1},
Yudong Wang\textsuperscript{\rm 2}$^{\dagger}$
} \\
\textsuperscript{\rm 1}ModelBest Inc. \quad
\textsuperscript{\rm 2}Tsinghua University ~~~~ \\
\texttt{pulushi@modelbest.cn} \quad
\texttt{yudongwang@tsinghua.edu.cn}
}

\newcommand\blfootnote[1]{%
\begingroup
\renewcommand\thefootnote{}\footnote{#1}%
\addtocounter{footnote}{-1}%
\endgroup
}

\begin{document}

\maketitle
\blfootnote{$\dagger$ Corresponding authors.}

\thispagestyle{fancy} 

\vspace{-2em}

\begin{abstract}
    Large Language Models (LLMs) have made notable progress in automated theorem proving, yet existing formal benchmarks remain limited in both mathematical coverage and difficulty. Most are concentrated in areas that are easier to formalize, such as algebra and elementary number theory, and provide limited coverage of subfields that require deeper reasoning, including mathematical analysis.
    To address this gap, we introduce \textit{\textbf{MA-ProofBench}}, to the best of our knowledge, the first formal theorem-proving benchmark dedicated to \textbf{M}athematical \textbf{A}nalysis.
    The benchmark contains 200 formalized theorems covering 6 core topics and 27 subcategories, including measure and integration theory, complex analysis, and functional analysis.
    The problems are divided into two difficulty levels, an undergraduate level (Level I, 100 problems) and a Ph.D. qualifying level (Level II, 100 problems), to evaluate how well LLMs perform formal reasoning at different mathematical depths.
    Each problem is constructed through a human-led, LLM-assisted formalization pipeline followed by independent expert review, ensuring that the formal statements remain faithful to the original mathematics.
    We evaluate a range of recent general-purpose reasoning models and formal theorem provers on MA-ProofBench.
    However, most models perform poorly: even the best-performing model, GPT-5.5, achieves only 16\% Pass@8 on Level I and 5\% on Level II, while most models stay close to 0\% on Level II.
    Further analysis identifies Mathlib hallucinations and incomplete proofs as the two dominant failure modes, while an evaluation on the natural-language version of the benchmark exposes a clear gap between informal and formal reasoning.
    MA-ProofBench is intended to serve as a reliable reference for tracking progress in formal mathematical reasoning in advanced domains.
    \end{abstract}
\section{Introduction}
The rapid advancement of Large Reasoning Models (LRMs)~\citep{jaech2024openai, guo2025deepseek} has driven significant progress in end-to-end automated theorem proving.
As natural language proofs become increasingly long and intricate, manual verification by human experts has emerged as a major bottleneck for AI-assisted mathematics.
Formal verification via interactive theorem provers such as Lean 4 \citep{moura2021lean} offers a scalable and reliable alternative.
Building on this approach, recent models \citep{lin2025goedelproverv2scalingformaltheorem, ren2025deepseekproverv2advancingformalmathematical, chen2025seed} have demonstrated the ability to generate non-trivial formal proofs across diverse benchmarks, including challenging problems from the International Mathematical Olympiad (IMO).

Despite this progress, existing formal benchmarks fall short of providing a comprehensive assessment of model capabilities in complex mathematical reasoning. The discriminative power of some traditional benchmarks is diminishing; for instance, Seed-Prover \citep{chen2025seed} has effectively saturated miniF2F \citep{zhengminif2f}, achieving a 100\% success rate.
Moreover, the quality of the formal statements directly affects the reliability of evaluation. Recent studies \citep{ammanamanchi2025faults, ospanov2026minifflean} have shown that some existing datasets contain semantic flaws, imprecise statements, or formalizations that do not fully match the intended mathematical meaning of the original problems.
More importantly, current benchmarks, such as FIMO~\citep{liu2023fimochallengeformaldataset}, Putnam~\citep{NEURIPS2024_1582eaf9}, ProofNet~\citep{azerbayev2023proofnetautoformalizingformallyproving}, and FormalMATH~\citep{yu2025formalmathbenchmarkingformalmathematical}, exhibit an uneven topic distribution. 
Many problems are concentrated in areas that are comparatively easier to formalize, such as algebra, elementary number theory, combinatorics, or discrete structures, while fields that require reasoning jointly about continuity, limits, and topological structures, such as measure theory, complex analysis, and functional analysis, remain underrepresented.

Mathematical Analysis (MA) is a central branch of modern mathematics, studying continuity, limits, and infinite processes.
Theorem proving in this domain is often demanding, as it requires both an understanding of the key structures involved and the ability to decompose a goal into a sequence of verifiable intermediate steps, making it a particularly challenging target for current formal systems.
To address the limited coverage of analysis in existing formal benchmarks, we introduce \textbf{\textit{MA-ProofBench}}, a high-quality, broad-coverage, two-tiered formal benchmark dedicated to mathematical analysis.

The problems in MA-ProofBench are mainly collected from widely used undergraduate analysis textbooks and publicly available Ph.D.\ qualifying exams, yielding two difficulty levels: \emph{Level~I} contains foundational exercises from the standard undergraduate curriculum, while \emph{Level~II} comprises more sophisticated analysis problems from doctoral qualifying examinations.
To ensure the mathematical fidelity of the formal statements, we employ a human-led, LLM-assisted formalization pipeline coupled with a rigorous independent expert review stage. The resulting dataset provides a robust framework for evaluating the formal reasoning capabilities of models in mathematical analysis.

We evaluate both general-purpose reasoning models and formal theorem provers on MA-ProofBench.
The evaluation reveals that both general-purpose reasoning models and formal theorem provers struggle with MA-ProofBench: the best-performing general-purpose reasoning model, GPT-5.5, achieves only 16\% Pass@8 on Level I and 5\% on Level II, while the strongest theorem prover, DeepSeek-Prover-V2-671B, reaches just 6.86\% and 0.44\%, respectively. 
These results show that current models still struggle with formal proofs in advanced analysis.
Further analysis attributes the majority of failures to Mathlib hallucinations and incomplete proofs, while an evaluation on the natural-language version of the benchmark exposes a clear gap between informal and formal reasoning.

Our main contributions are summarized as follows:
\begin{itemize}[leftmargin=*]
    \item We introduce \textbf{\textit{MA-ProofBench}}, the first formal benchmark dedicated to mathematical analysis. It comprises two difficulty levels (undergraduate and Ph.D.), each with 100 problems, covering 6 core topics and 27 subcategories. 
    \item We propose a human-led, LLM-assisted formalization workflow that addresses the semantic and syntactic challenges of translating advanced analysis problems into Lean 4, thereby ensuring the quality of the resulting formal statements.
    \item We evaluate a broad range of general-purpose reasoning models and formal theorem provers on MA-ProofBench. The results identify Mathlib hallucinations and incomplete proofs as the dominant failure modes, and further reveal a substantial gap between models' informal and formal reasoning abilities.
\end{itemize}

\section{Related Works}

\subsection{Formal Theorem Proving}
The field of automated theorem proving is evolving from single-paradigm search toward more sophisticated search-augmented and agent-based methods.
Since OpenAI's GPT-f \citep{polu2020generative} adopted best-first search and DeepMind's AlphaProof \citep{hubert2025olympiad} achieved strong results on IMO-style problems, the open-source community has explored several technical directions.
Tree-search-based methods such as InternLM2.5-StepProver \citep{wu2025internlm2}, BFS-Prover \citep{xin2025bfs}, DeepSeek-Prover-V1.5 \citep{xin2024deepseekproverv15harnessingproofassistant}, and HunyuanProver \citep{li2024hunyuanprover} explore tactic-level tree search algorithms such as Monte Carlo Tree Search.
In contrast, whole-proof generation methods such as Kimina-Prover \citep{wang2025kimina}, DeepSeek-Prover-V2 \citep{ren2025deepseekproverv2advancingformalmathematical}, and Goedel-Prover \citep{lin2025goedelproverv2scalingformaltheorem} generate the entire proof in a single pass.
Another line of work augments formal proof search and verification by building agent-based systems, including Seed-Prover \citep{chen2025seed, chen2025seed15}, Aristotle \citep{achim2025aristotle}, Ax-Prover \citep{breen2025axproverdeepreasoningagentic}, and Numina-Lean-Agent \citep{liu2026numina}.
\subsection{Mathematical Benchmarks}
Existing mathematical benchmarks can be broadly categorized as informal or formal.
Informal benchmarks including GSM8K \citep{cobbe2021trainingverifierssolvemath}, MATH \citep{hendrycks2021measuring}, AIME \citep{aime}, OlympiadBench \citep{he-etal-2024-olympiadbench}, Omni-MATH \citep{gao2024omnimathuniversalolympiadlevel}, IMO-AnswerBench \citep{luong-etal-2025-towards}, and AMO-Bench \citep{an2025amobenchlargelanguagemodels} focus mainly on numerical problem solving with final-answer evaluation, while IMO-ProofBench \citep{luong-etal-2025-towards} and ProofBench \citep{ma2025reliable} evaluate natural language proof generation.
In formal mathematics, early benchmarks such as miniF2F \citep{zhengminif2f} and FIMO \citep{liu2023fimochallengeformaldataset} primarily cover competition-level problems, ranging from high school competitions to the IMO. 
Subsequent benchmarks such as ProofNet \citep{azerbayev2023proofnetautoformalizingformallyproving}, PutnamBench \citep{NEURIPS2024_1582eaf9}, and FormalMATH \citep{yu2025formalmathbenchmarkingformalmathematical} expand the scope to the undergraduate level, but their content remains concentrated in areas such as algebra, topology, and elementary calculus.
Recent benchmarks also target specific domains, such as CombiBench \citep{liu2025combibench}, FATE \citep{jiang2026fate}, and LeanCat \citep{xu2025leancat}. 
MA-ProofBench follows this domain-specific line of work and focuses on mathematical analysis, a subfield that remains underrepresented in existing formal benchmarks.
\section{MA-ProofBench Construction and Characteristics}

\subsection{Benchmark Overview}

\begin{wraptable}{r}{0.58\textwidth} 
    \centering
    \vspace{-1.5em} 
    \caption{Difficulty Level Distribution}
    \label{tab:difficulty_wrap}
    \small 
    \begin{tabularx}{\linewidth}{l l X c} 
        \toprule
        \textbf{Level} & \textbf{Description} & \textbf{Source} & \textbf{Count} \\
        \midrule
        Level I  & Undergraduate & Basic Textbook Exercises & 100 \\
        \addlinespace[0.5em]
        Level II & Ph.D. & Exam Problems From Top-Tier Universities & 100 \\
        \bottomrule
    \end{tabularx}
    \vspace{-1em} 
\end{wraptable}

\begin{figure}[htbp]
    \centering
    \begin{subfigure}[b]{0.48\textwidth}
        \centering
        \includegraphics[width=\textwidth]{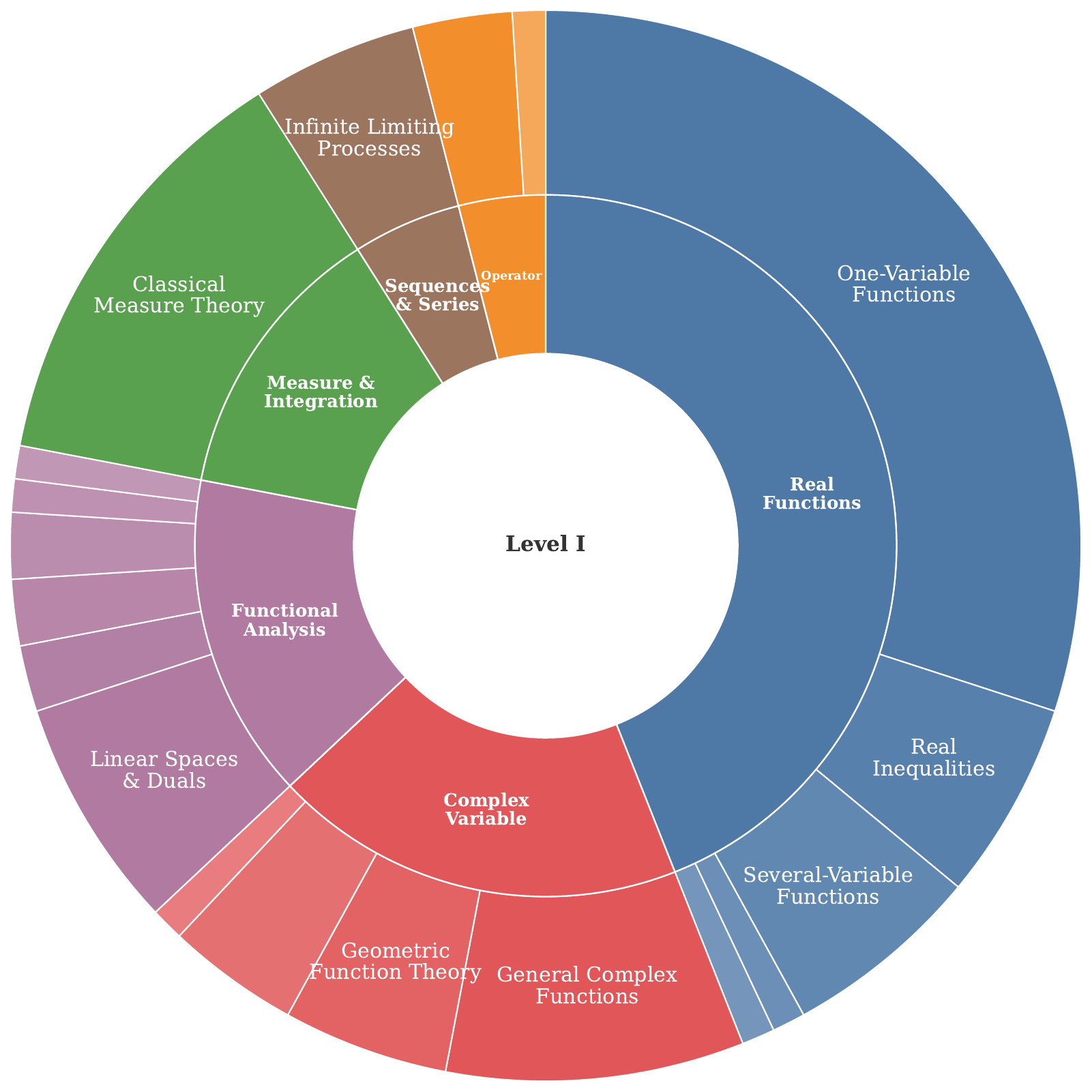}
        \caption{Category distribution for Level I}
        \label{fig:donut_level1}
    \end{subfigure}
    \hfill 
    \begin{subfigure}[b]{0.48\textwidth}
        \centering
        \includegraphics[width=\textwidth]{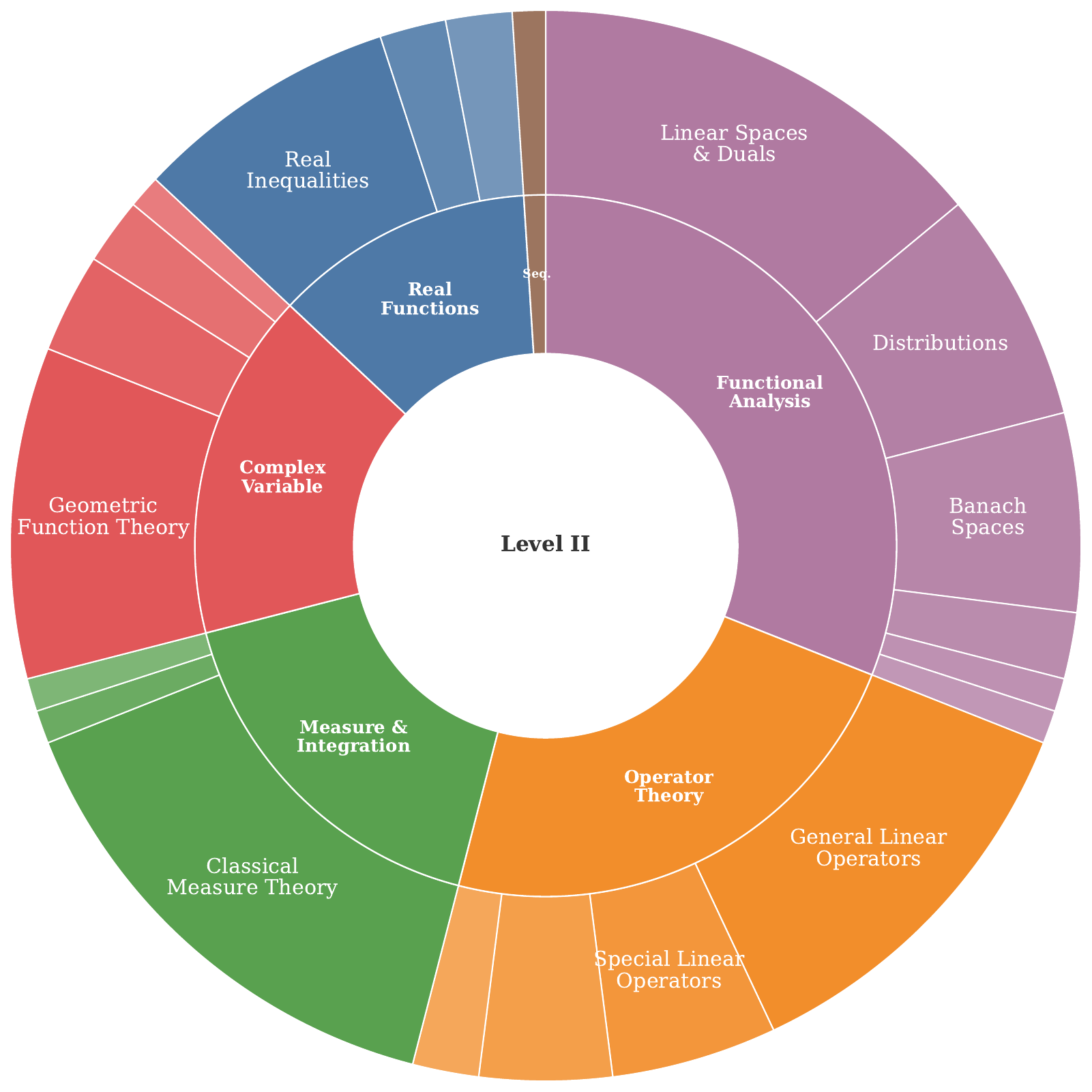}
        \caption{Category distribution for Level II}
        \label{fig:donut_level2}
    \end{subfigure}
    
    \caption{
    Category distribution of MA-ProofBench across Level I and Level II problems.
    The inner ring represents high-level mathematical topics, while the outer ring shows their finer-grained subcategories.
    For readability, only subcategories with relatively high frequency are annotated in the outer ring, although all subcategories are included in the proportional areas.
    A detailed breakdown of the category distribution is provided in Appendix \ref{topic_class}.
    }    
    \label{fig:distribution_donut}
\end{figure}

The problems in MA-ProofBench are collected primarily from widely used undergraduate textbooks in mathematical analysis and publicly accessible Ph.D. qualifying examination papers. 
Specifically, the problems are organized according to the Mathematics Subject Classification (MSC), encompassing 6 core categories of analysis: Real Functions, Measure and Integration, Functions of a Complex Variable, Sequences, Series, and Summability, Functional Analysis, and Operator Theory. These are further subdivided into 27 subcategories, with the detailed distribution illustrated in Table \ref{tab:difficulty_wrap} and Figure \ref{fig:distribution_donut}.

\begin{figure}[htbp]
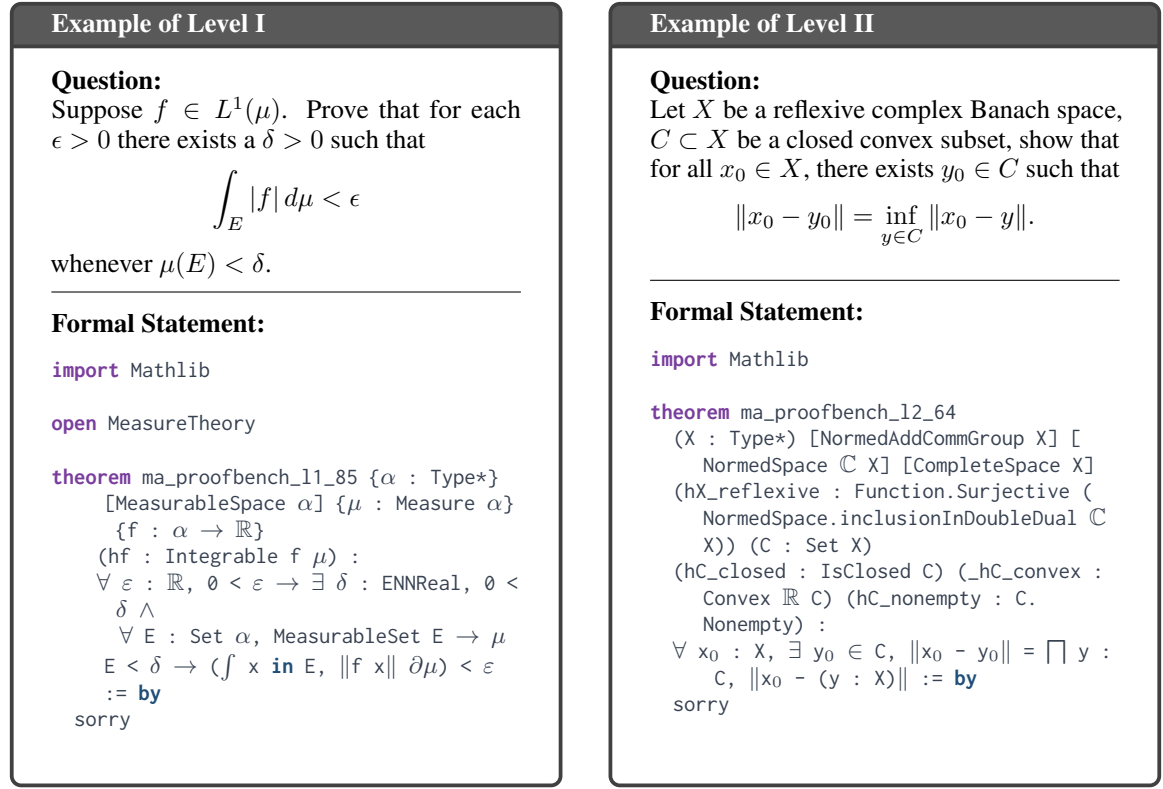

    \centering

    \begin{minipage}[t]{0.48\textwidth}
        \begin{tcolorbox}[
            enhanced,
            fonttitle=\bfseries,
            title={Example of Level I},
            colback=white,
            colframe=black!75,
            colbacktitle=black!65,
            coltitle=white,
            equal height group=level_examples_grp,
        ]
        \textbf{Question:} \\
        Suppose $f\in L^1(\mu)$. Prove that for each $\epsilon>0$ there exists a $\delta>0$ such that
        \[
        \int_E |f|\,d\mu<\epsilon
        \]
        whenever $\mu(E)<\delta$.

        \vspace{0.5em}
        \hrule
        \vspace{0.8em}
        \textbf{Formal Statement:}\\[-0.4em]
        \begin{lstlisting}[language=lean4, aboveskip=0pt, belowskip=0pt]
import Mathlib

open MeasureTheory

theorem ma_proofbench_l1_85 {α : Type*} [MeasurableSpace α] {μ : Measure α} {f : α → ℝ}
    (hf : Integrable f μ) :
    ∀ ε : ℝ, 0 < ε → ∃ δ : ENNReal, 0 < δ ∧
      ∀ E : Set α, MeasurableSet E → μ E < δ → (∫ x in E, ‖f x‖ ∂μ) < ε := by
  sorry
        \end{lstlisting}
        \end{tcolorbox}
    \end{minipage}%
    \hfill
    \begin{minipage}[t]{0.48\textwidth}
        \begin{tcolorbox}[
            enhanced,
            fonttitle=\bfseries,
            title={Example of Level II},
            colback=white,
            colframe=black!75,
            colbacktitle=black!65,
            coltitle=white,
            equal height group=level_examples_grp,
        ]
        \textbf{Question:} \\
        Let $X$ be a reflexive complex Banach space, $C\subset X$ be a closed convex subset, show that for all $x_0\in X$, there exists $y_0\in C$ such that
        \[
        \|x_0-y_0\|=\inf_{y\in C}\|x_0-y\|.
        \]

        \vspace{0.5em}
        \hrule
        \vspace{0.8em}
        \textbf{Formal Statement:}\\[-0.4em]
        \begin{lstlisting}[language=lean4, aboveskip=0pt, belowskip=0pt]
import Mathlib

theorem ma_proofbench_l2_64
  (X : Type*) [NormedAddCommGroup X] [NormedSpace ℂ X] [CompleteSpace X]
  (hX_reflexive : Function.Surjective (NormedSpace.inclusionInDoubleDual ℂ X)) (C : Set X)
  (hC_closed : IsClosed C) (_hC_convex : Convex ℝ C) (hC_nonempty : C.Nonempty) :
  ∀ x₀ : X, ∃ y₀ ∈ C, ‖x₀ - y₀‖ = ⨅ y : C, ‖x₀ - (y : X)‖ := by
  sorry
        \end{lstlisting}
        \end{tcolorbox}
    \end{minipage}

    \caption{Comparison of problem difficulty. The Level I example involves a basic absolute continuity theorem in measure theory with straightforward typing. In contrast, the Level II example requires setting up heavy mathematical machinery, such as reflexive Banach spaces and double duals, illustrating the increased conceptual and syntactic complexity.}
    \label{fig:level_examples}
\end{figure}

In terms of topic distribution, Level I primarily focuses on foundational topics in analysis, such as functions of one variable and classical measure theory. 
In contrast, Level II emphasizes deeper abstract structures, with a strong focus on advanced topics such as linear function spaces in functional analysis, the general theory of linear operators, advanced measure theory, and geometric function theory. Figure~\ref{fig:level_examples} shows representative examples for each level.

\subsection{Curation Workflow}

\begin{figure}[htbp]
    \centering
    \includegraphics[width=1.0\linewidth]{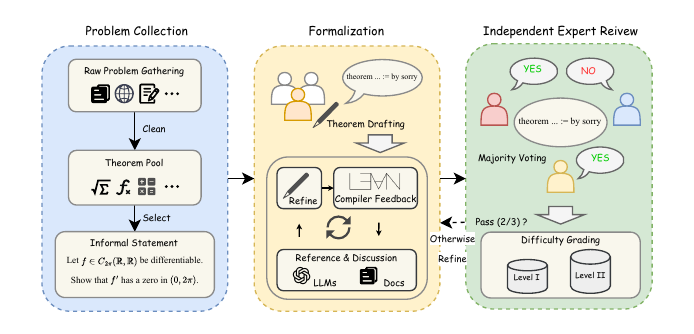}
    \caption{Overview of the curation workflow of MA-ProofBench, comprising Problem Collection, Formalization, Independent Expert Review, and Difficulty Grading.}
    \label{fig:workflow}
\end{figure}

We constructed the benchmark through a four-stage workflow driven by human-LLM collaboration, as illustrated in Figure \ref{fig:workflow}:

\begin{enumerate}[leftmargin=*]
    \item \textbf{Problem Collection:} We collected raw problems and conducted several rounds of cleaning and proofreading to correct inaccurate statements, standardize the mathematical content, and unify the formatting, removing ambiguous or incomplete entries along the way. This yielded a candidate pool of about 500 problems, from which we selected a representative subset of 200 problems.

    \item \textbf{Formalization:} Human experts first translated each natural-language problem into a draft Lean 4 statement, leaving the proof as \texttt{sorry}. They then iterated in a ``formalize--feedback--refine'' loop, using Lean 4 compiler messages (e.g., type mismatches or unknown identifiers) to fix the code and add missing imports. For unfamiliar Mathlib APIs, experts consulted LLM suggestions and cross-checked them against the official Mathlib documentation. Final decisions were always made by the human experts, ensuring that the formal statements remained faithful to the original problems.

    \item \textbf{Independent Expert Review:} Each formal statement was then independently reviewed by three additional experts. Each reviewer was required to ``reverse-translate'' the Lean 4 code back into its intended mathematical statement, ensuring that no implicit assumption was dropped and no condition was inadvertently weakened. A theorem was accepted only if at least two reviewers voted in favor; otherwise, it was sent back to Stage 2 for revision.

    \item \textbf{Difficulty Grading:} Based on the intrinsic mathematical complexity and the difficulty of formalization, experts scored and categorized the problems into Level I and Level II. The evaluation criteria include:
    \begin{itemize}[leftmargin=*]
        \item The number of critical sub-tasks or lemmas required for the proof.
        \item The level of conceptual abstraction (e.g., elementary limit calculations versus the construction of abstract measure spaces).
    \end{itemize}
\end{enumerate}

\subsection{Formalization Standards} %
To ensure the semantic fidelity of each formal statement to its informal counterpart, MA-ProofBench adheres to a unified set of formalization conventions:

\noindent\textbf{Function Definitions.}
In informal mathematics, a function is sometimes defined only on a certain subset. In formalization, however, we typically represent it as a total function on the ambient space and make the actual domain of interest explicit through a predicate or a set condition. This treatment keeps the function type uniform and avoids frequent coercions between subtypes and the ambient space, thereby allowing us to directly reuse Mathlib's general definitions and theorems for notions such as continuity, differentiability, and harmonicity. In this setting, the values outside the specified subset do not carry mathematical content; the relevant statements are asserted only on the prescribed set or in its neighborhood.

\noindent\textbf{Integration Framework.}
For integrals in natural-language problem statements whose values lie in a normed linear space, in particular for real- or complex-valued integrals, we formalize them as Bochner integrals in Lean. This convention applies only when the target space carries the relevant normed linear structure; for objects such as extended nonnegative real-valued integrals, we still use Mathlib's dedicated framework for nonnegative Lebesgue integration.

\noindent\textbf{Spatial Structures.}
Natural-language problem statements sometimes leave implicit the distinction between $L^p$ equivalence classes and pointwise-defined functions, whereas formalization requires this distinction to be made explicit. We therefore choose the representation according to the mathematical structure on which each statement depends. For properties intrinsic to $L^p$ spaces, such as norms, distances, convergence, and bounded linear operators, functions are formalized as equivalence classes modulo almost-everywhere equality, which preserves the standard normed space structure of $L^p$. For statements involving values at specific points, pointwise inequalities, or pointwise regularity, we instead work with concrete function representatives.

\noindent\textbf{Explicit Premises.}
Implicit mathematical assumptions are stated explicitly in the formal theorem. Because Lean quantifies over every object satisfying the declared premises, non-degeneracy assumptions that are commonly left tacit in informal writing, such as nonemptiness captured by \texttt{[Nonempty X]}, are added directly to the statement. Doing so prevents the formal theorem from being trivially or vacuously satisfied in boundary or degenerate cases, and keeps its semantic scope aligned with that of the original problem.

\section{Experiments}
\subsection{Experimental Setting}

We evaluated a range of models on MA-ProofBench, including formal theorem provers (DeepSeek-Prover-V2 (7B and 671B) \citep{ren2025deepseekproverv2advancingformalmathematical}, Kimina-Prover-72B \citep{wang2025kimina}, and Goedel-Prover-V2 (8B and 32B) \citep{lin2025goedelproverv2scalingformaltheorem}); open-source general-purpose reasoning models, including DeepSeek-V3.2-Thinking \citep{deepseekai2025deepseekv32pushingfrontieropen}, GLM-5.1 \citep{zai2026glm5}, Qwen3.5-397B-A17B \citep{qwen3.5}, Qwen3-235B-Thinking-2507 \citep{qwen3technicalreport}, Nemotron-3-Nano-30B-A3B \citep{nvidia_nemotron_nano_v3_2025}, and GPT-OSS-120B (High) \citep{openai2025gptoss}; and proprietary models, namely GPT-5.5 (xhigh) \citep{openai2026gpt55}, Gemini 3.1 Pro (High) \citep{deepmind2026gemini31pro}, and Claude Sonnet 4.6 (High) \citep{anthropic2026sonnet46news}.
The generation parameters were configured with a maximum output length of 32k tokens and a temperature of 1.0. 
For evaluation, we used the Pass@$k$ metric \citep{chen2021evaluatinglargelanguagemodels}, which is formally defined as follows:
\[
\text{Pass}@k = \frac{1}{N} \sum_{i=1}^{N} \left( 1 - \frac{\binom{n-c_i}{k}}{\binom{n}{k}} \right),
\]
where $N$ is the total number of problems in the benchmark, $n$ denotes the number of candidate proofs generated per problem, and $c_i$ denotes the number of successful proofs among the $n$ candidates for the $i$-th problem.

For all open-source models, we generated $n=32$ candidate proofs per problem, while for the proprietary GPT-5.5 (xhigh), Gemini 3.1 Pro (High), and Claude Sonnet 4.6 (High), we used $n=8$ due to API budget constraints. We therefore adopt Pass@8 as the primary metric for cross-model comparison, and report Pass@32 only as a supplementary result for the models sampled with $n=32$. 
All evaluations were run on Mathlib 4.28.0 \citep{10.1145/3372885.3373824}, with Kimina Lean Server \citep{santos2025kiminaleanserverhighperformance} as the compilation backend for its efficiency.
We establish the following criteria for a proof to be considered successful:
\begin{itemize}[leftmargin=*]
    \item The proof must pass compiler verification without generating any errors.
    \item The proof must not contain any \code{sorry} placeholders.
    \item The proof must include all prerequisite components (e.g., definitions) required by the original problem, while leaving the theorem statement itself strictly unaltered.
\end{itemize}
The exact prompts used in our evaluation are provided in Appendix \ref{appendix_prompts}.

\subsection{Main Results}

\begin{table*}[h]
    \centering
    \caption{Performance evaluation of different models on MA-ProofBench. Accuracies (\%) are reported under Pass@1, Pass@8, and Pass@32 settings. Pass@8 is used as the primary cross-model comparison; Pass@32 is reported only for the models sampled with $n=32$.}
    \label{tab:main_results}
    \small
    \begin{tabular}{l ccc ccc}
    \toprule
    \multirow{2}{*}{\textbf{Model}} & \multicolumn{3}{c}{\textbf{Level I}} & \multicolumn{3}{c}{\textbf{Level II}} \\
    \cmidrule(lr){2-4} \cmidrule(lr){5-7}
    & Pass@1 & Pass@8 & Pass@32 & Pass@1 & Pass@8 & Pass@32 \\
    \midrule
    
    \rowcolor[gray]{0.95} \multicolumn{7}{l}{\textit{\textbf{Closed-Source Reasoning Models}}} \\
    GPT-5.5 (xhigh) & 6.50 & 16.00 & -- & 1.75 & 5.00 & -- \\
    Gemini 3.1 Pro (High) & 4.38 & 13.00 & -- & 1.63 & 5.00 & -- \\
    Claude Sonnet 4.6 (High) & 3.50 & 6.00 & -- & 0.50 & 3.00 & -- \\
    \midrule
    
    \rowcolor[gray]{0.95} \multicolumn{7}{l}{\textit{\textbf{Open-Source Reasoning Models}}} \\
    DeepSeek-V3.2-Thinking & 2.09 & 5.56 & 7.00 & 0.66 & 1.85 & 2.00 \\
    GLM-5.1 & 0.06 & 0.44 & 1.00 & 0.00 & 0.00 & 0.00 \\
    Qwen3.5-397B-A17B & 0.87 & 3.14 & 5.00 & 0.00 & 0.00 & 0.00 \\
    Qwen3-235B-Thinking-2507 & 0.03 & 0.25 & 1.00 & 0.00 & 0.00 & 0.00 \\
    Nemotron-3-Nano-30B-A3B & 2.12 & 3.59 & 4.00 & 0.00 & 0.00 & 0.00 \\
    GPT-OSS-120B (High) & 0.06 & 0.50 & 2.00 & 0.00 & 0.00 & 0.00 \\
    \midrule
    
    \rowcolor[gray]{0.95} \multicolumn{7}{l}{\textit{\textbf{Theorem Provers}}} \\
    DeepSeek-Prover-V2-671B & 3.22 & 6.86 & 9.00 & 0.06 & 0.44 & 1.00 \\
    Kimina-Prover-72B & 1.12 & 3.11 & 4.00 & 0.00 & 0.00 & 0.00 \\
    Goedel-Prover-V2-32B & 2.84 & 5.05 & 6.00 & 0.00 & 0.00 & 0.00 \\
    Goedel-Prover-V2-8B & 1.97 & 3.84 & 5.00 & 0.00 & 0.00 & 0.00 \\
    DeepSeek-Prover-V2-7B & 0.94 & 3.46 & 6.00 & 0.00 & 0.00 & 0.00 \\
    \bottomrule
    \end{tabular}
\end{table*}

Table~\ref{tab:main_results} summarizes the main evaluation results, from which we highlight three observations.
\begin{itemize}[leftmargin=*]
    \item \textbf{Closed-source reasoning models lead overall.} GPT-5.5 (xhigh) attains the highest Pass@8 on Level I (16\%) and ties Gemini 3.1 Pro (High) for the best Pass@8 on Level II (5\%).
    \item \textbf{Formal theorem provers remain strong on Level I.} DeepSeek-Prover-V2-671B reaches 6.86\% Pass@8 and 9\% Pass@32 on Level I, outperforming all open-source general-purpose reasoning models on this split. A separate observation is that, despite having only $\sim$3B active parameters, Nemotron-3-Nano-30B-A3B reaches 2.12\% Pass@1 and 3.59\% Pass@8 on Level I, clearly above several open-source reasoning models of similar or larger scale. Given that its post-training stage incorporates large-scale Lean~4 formal-proof SFT examples, this result suggests that formal-proof data may help smaller models acquire direct proof-generation ability. On Level II, however, formal theorem provers reach at most 1\% Pass@32 and no longer outperform the best open-source general-purpose reasoning model, DeepSeek-V3.2-Thinking, which reaches 2\%. This indicates that the theorem prover advantage observed on Level I does not carry over to problems requiring deeper analytical reasoning.
    \item \textbf{Level II exposes a substantial capability gap.} All model families perform much worse on Level II, and even the strongest models solve only a small fraction of the problems. This suggests that this split stresses capabilities beyond local tactic selection or short proof completion, including long-range proof planning, the construction of domain-specific auxiliary lemmas, and reliable use of deeper Mathlib knowledge in mathematical analysis.
\end{itemize}

\section{Further Analysis} \label{sec:analysis}
\subsection{Error Classification in Formal Proofs}

\begin{figure}[htbp]
    \centering
    \begin{subfigure}[b]{0.48\textwidth}
        \centering
        \includegraphics[width=\textwidth]{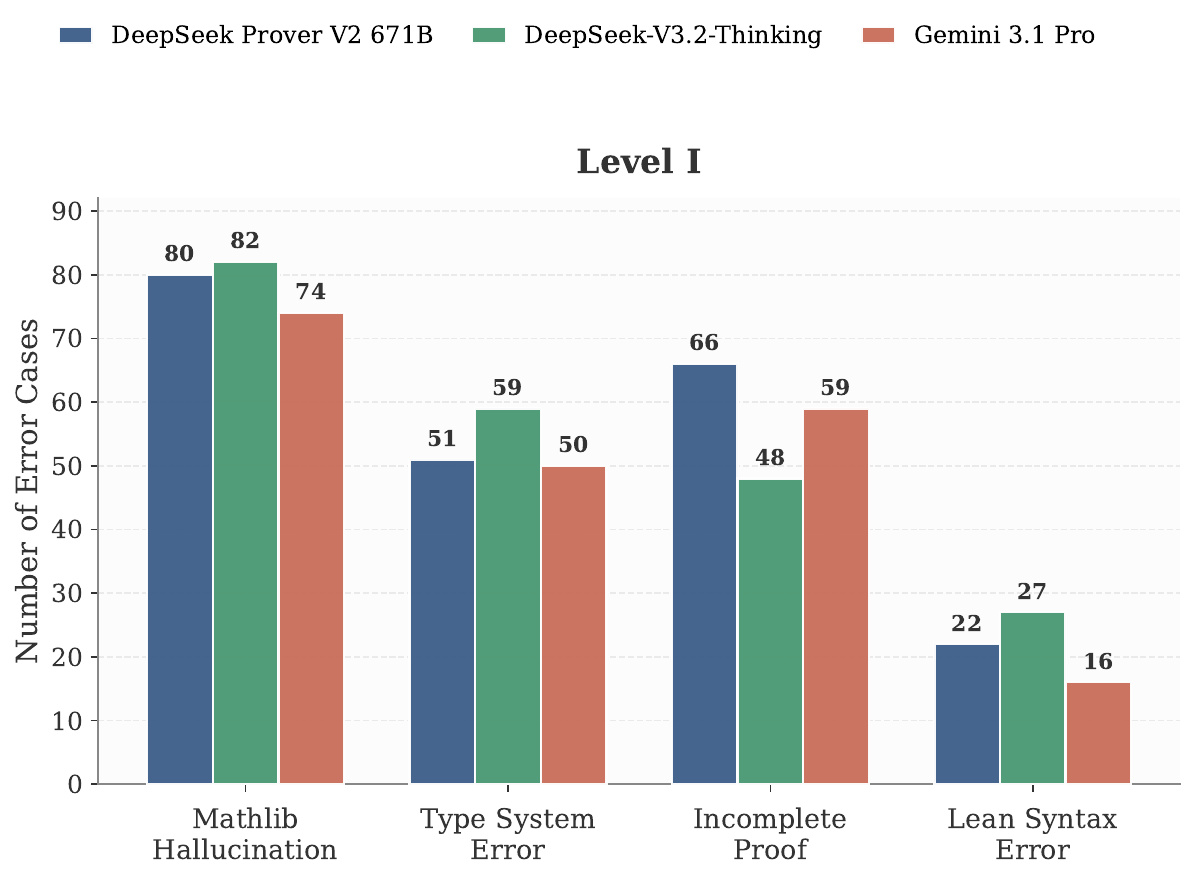}
        \caption{Error category distribution on Level I}
        \label{fig:error_level1}
    \end{subfigure}
    \hfill 
    \begin{subfigure}[b]{0.48\textwidth}
        \centering
        \includegraphics[width=\textwidth]{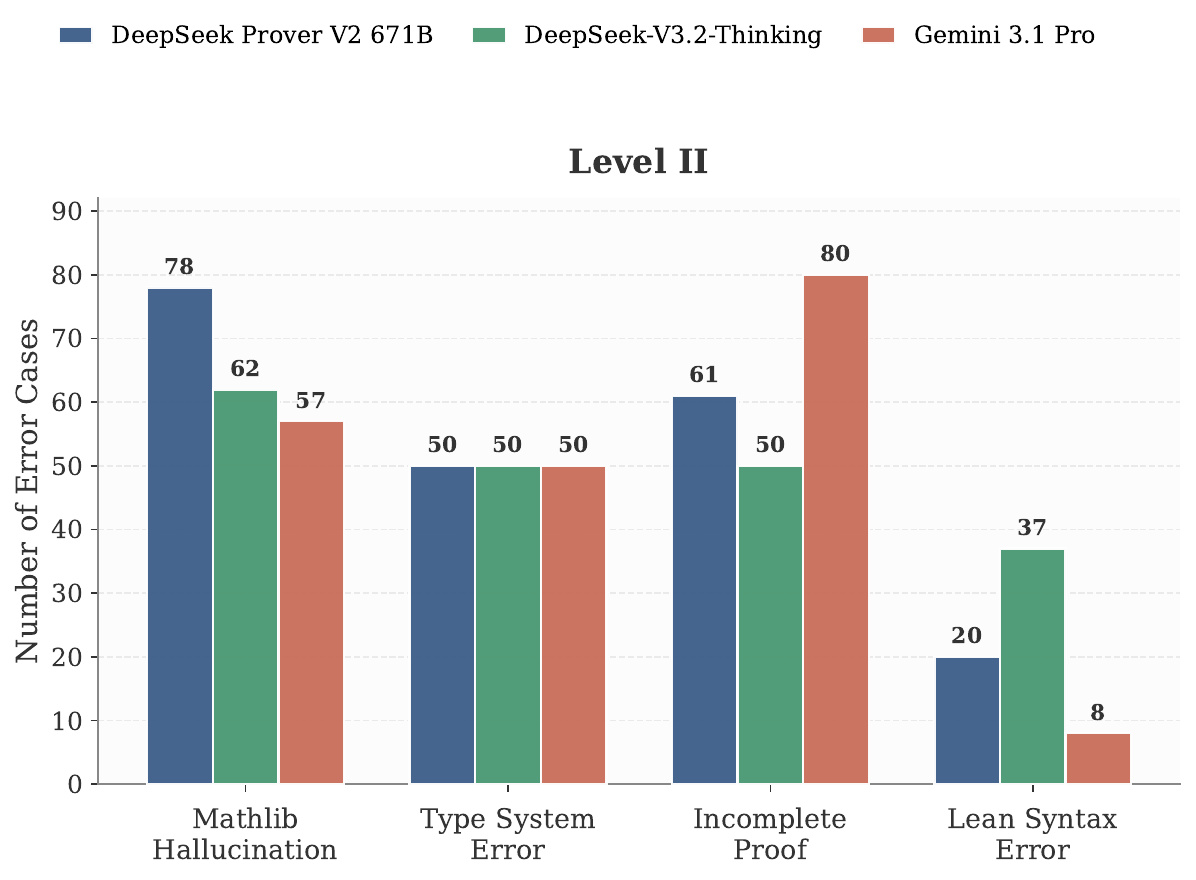}
        \caption{Error category distribution on Level II}
        \label{fig:error_level2}
    \end{subfigure}
    
    \caption{
    Distribution of error categories across representative models on MA-ProofBench.
    The figure reports the number of failed proof attempts assigned to each error type, including Mathlib hallucination, type system error, incomplete proof, and Lean syntax error.
    }    
    \label{fig:error_distribution}
\end{figure}

To better understand why models fail, we perform an error attribution analysis based on Lean compiler feedback and manual inspection, and summarize the failed proofs into four major error types.

\noindent\textbf{Mathlib Hallucinations.}
These errors stem from the model referencing fictitious or mismatched Mathlib definitions, theorems, and identifiers, which trigger \texttt{Unknown Constant} or \texttt{Unknown Identifier} compiler errors.
Two common cases are \textit{namespace omission}, where the referenced theorem exists in Mathlib but the model fails to \texttt{open} the corresponding namespace, and \textit{name fabrication}, where the model ``guesses'' and fabricates non-existent theorem names based on Mathlib's naming conventions in an attempt to complete the proof.

\noindent\textbf{Type System Errors.}
The type of an expression does not match the type expected in context.
In analysis, the most common case is confusing $\mathbb{R}$ with \texttt{ENNReal} (extended non-negative reals), which suggests that the model does not clearly separate metric values from measure values.
This category also covers mismatched function signatures and typeclass synthesis failures, where the model cannot supply the structures required for the proof (e.g., \texttt{MetricSpace} or \texttt{MeasurableSpace}).

\noindent\textbf{Incomplete Proofs.}
These proofs contain unresolved sub-goals or \code{sorry} placeholders.
Typical patterns include \textit{tactic misuse}, where the model employs general-purpose tactics for tasks beyond their capabilities, such as using \texttt{linarith} for complex non-linear inequalities or expecting \texttt{simp} to automate deep analytical reasoning, and \textit{explicit sorry insertion}, where the model deliberately inserts \code{sorry} to bypass substantial logical leaps or complex sub-proofs.

\noindent\textbf{Lean Syntax Errors.}
This category refers to code that violates Lean 4 syntax rules, causing the parser to fail.
Typical phenomena include \textit{mode confusion}, where the model fails to distinguish the syntactic boundaries between Term Mode and Tactic Mode, and \textit{illegal identifiers}, such as the misuse of reserved words or special symbols (e.g., using $\lambda$ in inappropriate contexts), or code structures that do not conform to the Lean 4 macro system specifications.

Furthermore, we conducted a random sampling of the failure cases produced by DeepSeek-Prover-V2-671B, DeepSeek-V3.2-Thinking, and Gemini 3.1 Pro across both Level I and Level II problems, with the resulting error distribution summarized in Figure \ref{fig:error_distribution}.
Overall, Mathlib hallucinations and incomplete proofs constitute the primary sources of error. This trend is particularly pronounced for DeepSeek-Prover-V2-671B and DeepSeek-V3.2-Thinking, which yielded 80 and 82 such instances in Level I, and 78 and 62 instances in Level II, respectively.
Gemini 3.1 Pro shows a different pattern on Level II: its dominant failure mode is incomplete proofs (80 instances), much higher than the other models, suggesting that it tends to leave proofs unfinished on harder problems.
Type system errors occur at a fairly stable rate across models and difficulty levels.
Although Lean syntax errors are the least frequent overall, DeepSeek-V3.2-Thinking has more of these errors on Level II (37 instances), suggesting lower syntactic stability on harder problems.

\subsection{Gap Between Informal and Formal Reasoning}

An inspection of the reasoning traces shows that, on most problems, the models follow a two-stage strategy: an informal proof is first developed within the chain-of-thought, and only then translated into Lean 4 code.
Under this pipeline, our main results reveal a phenomenon worth highlighting: several open-source models with otherwise strong general reasoning ability, such as GLM-5.1 and Qwen3-235B-Thinking-2507, score close to zero in the formal setting. A natural question is whether this gap originates from a deficiency in the underlying mathematical reasoning, or from the difficulty of converting an informal proof into Lean 4.

\begin{wraptable}{r}{0.5\textwidth}
    \centering
    \vspace{-1.2em}
    \caption{Number of fully correct Pass@1 informal proofs per level, as judged by GPT-5.5.}
    \label{tab:informal_results}
    \small
    \setlength{\tabcolsep}{6pt}
    \begin{tabular}{l c c}
        \toprule
        \textbf{Model} & \textbf{Level I} & \textbf{Level II} \\
        \midrule
        Qwen3-235B-Thinking-2507 & 66 & 42 \\
        DeepSeek-V3.2-Thinking   & 85 & 66 \\
        GLM-5.1                  & 90 & 75 \\
        \bottomrule
    \end{tabular}
    \vspace{-1em}
\end{wraptable}

To address this question, we evaluate the same models on the natural-language version of MA-ProofBench. Each model produces a single informal proof per problem under Pass@1, which is then graded by GPT-5.5 on a three-level scale (0, 0.5, or 1) following the protocol of DeepSeekMath-V2~\citep{shao2025deepseekmathv2}. Table~\ref{tab:informal_results} reports the number of fully correct proofs that received a score of 1, the judging prompt is given in Appendix~\ref{appendix_prompts}.

Juxtaposing these results with the formal evaluation in Table~\ref{tab:main_results} reveals a clear gap between informal and formal reasoning: GLM-5.1 produces 90 and 75 fully correct informal proofs across the two levels, while its Pass@32 in the formal setting remains close to zero, and the same pattern is observed for Qwen3-235B-Thinking-2507.
For these models, the bottleneck lies in faithfully expressing such reasoning as Lean 4 code that the compiler accepts.

At the same time, formal performance is not fully determined by informal proficiency. DeepSeek-V3.2-Thinking scores below GLM-5.1 on the informal task, yet remains the strongest open-source general-purpose reasoning model in the formal setting, indicating that familiarity with Mathlib, handling of the type system, and the ability to discharge remaining subgoals all contribute substantially to the final formal score.

\section{Conclusion}

In this paper, we introduce \textbf{\textit{MA-ProofBench}}, a formal theorem-proving benchmark dedicated to mathematical analysis, comprising 200 theorems across two difficulty levels (Level~I and 
Level~II). 
Our human-led, LLM-assisted formalization pipeline, combined with independent expert review, ensures broad coverage of analytical subfields, high-quality formal statements, and substantial difficulty for current systems.
Both general-purpose reasoning models and formal theorem provers struggle with MA-ProofBench, with Mathlib hallucinations and incomplete proofs as the main bottlenecks, while an evaluation on the natural-language version of the benchmark further exposes a substantial translation gap between informal reasoning and verifiable Lean 4 proofs.
These findings suggest that future systems will require better grounding in Mathlib, stronger proof-completion capabilities for analysis-heavy Lean proofs, and more reliable translation from informal proofs into compilable Lean 4 code. We hope MA-ProofBench will serve as a rigorous testbed for advancing formal reasoning in advanced mathematics.


\bibliographystyle{citation}
\bibliography{citation}

\clearpage
\appendix
\section{Category Distribution} \label{topic_class}

MA-ProofBench covers a wide range of analytical topics. To illustrate its mathematical diversity, Table \ref{tab:msc_distribution} details the distribution of the benchmark's problems across various Mathematics Subject Classification (MSC) categories for both difficulty levels.

\begin{table*}[h]
\centering
\caption{Distribution of MA-ProofBench problems across MSC categories.}
\label{tab:msc_distribution}
\small
\renewcommand{\arraystretch}{1.08}
\setlength{\tabcolsep}{6pt}

\begin{tabularx}{\textwidth}{
  >{\raggedright\arraybackslash}X
  >{\centering\arraybackslash}p{1.4cm}
  >{\centering\arraybackslash}p{1.4cm}
}
\toprule
\textbf{Category / Subcategory Name} & \textbf{Level I} & \textbf{Level II} \\
\midrule

\rowcolor[gray]{0.95}
\multicolumn{3}{l}{\textbf{\textit{Real Functions}}} \\
\quad Functions of One Variable & 30 & 2 \\
\quad Functions of Several Variables & 6 & 2 \\
\quad Polynomials, Rational Functions in Real Analysis & 1 & 0 \\
\quad Inequalities in Real Analysis & 6 & 8 \\
\quad Miscellaneous Topics in Real Functions & 1 & 0 \\
\addlinespace[2pt]

\rowcolor[gray]{0.95}
\multicolumn{3}{l}{\textbf{\textit{Measure and Integration}}} \\
\quad Classical Measure Theory & 13 & 15 \\
\quad Set Functions, Measures and Integrals with Values in Abstract Spaces & 0 & 1 \\
\quad Set Functions and Measures on Spaces with Additional Structure & 0 & 1 \\
\addlinespace[2pt]

\rowcolor[gray]{0.95}
\multicolumn{3}{l}{\textbf{\textit{Functions of a Complex Variable}}} \\
\quad General Properties of Functions of One Complex Variable & 9 & 2 \\
\quad Series Expansions of Functions of One Complex Variable & 1 & 0 \\
\quad Geometric Function Theory & 5 & 10 \\
\quad Entire and Meromorphic Functions of One Complex Variable, and Related Topics & 4 & 3 \\
\quad Miscellaneous Topics of Analysis in the Complex Plane & 0 & 1 \\
\addlinespace[2pt]

\rowcolor[gray]{0.95}
\multicolumn{3}{l}{\textbf{\textit{Sequences, Series, and Summability}}} \\
\quad Convergence and Divergence of Infinite Limiting Processes & 5 & 1 \\
\addlinespace[2pt]

\rowcolor[gray]{0.95}
\multicolumn{3}{l}{\textbf{\textit{Functional Analysis}}} \\
\quad Topological Linear Spaces and Related Structures & 1 & 2 \\
\quad Normed Linear Spaces and Banach Spaces; Banach Lattices & 2 & 6 \\
\quad Inner Product Spaces and Their Generalizations, Hilbert Spaces & 2 & 0 \\
\quad Linear Function Spaces and Their Duals & 7 & 14 \\
\quad Distributions, Generalized Functions, Distribution Spaces & 0 & 7 \\
\quad Measures, Integration, Derivative, Holomorphy & 2 & 1 \\
\quad Commutative Banach Algebras and Commutative Topological Algebras & 0 & 1 \\
\quad Nonlinear Functional Analysis & 1 & 0 \\
\addlinespace[2pt]

\rowcolor[gray]{0.95}
\multicolumn{3}{l}{\textbf{\textit{Operator Theory}}} \\
\quad General Theory of Linear Operators & 3 & 12 \\
\quad Special Classes of Linear Operators & 0 & 5 \\
\quad Ordinary Differential Operators & 0 & 2 \\
\quad Integral, Integro-Differential, and Pseudodifferential Operators & 0 & 4 \\
\quad Nonlinear Operators and Their Properties & 1 & 0 \\
\midrule

\rowcolor[gray]{0.90}
\textbf{Total} & \textbf{100} & \textbf{100} \\
\bottomrule
\end{tabularx}
\end{table*}
\section{Case Studies}
\subsection{Successful Proofs}





\begin{examplebox}[app:correct_answer_2]{Success Case (DeepSeek-V3.2-Thinking)}
\textbf{Question:} \\
Let $f\in C_{2\pi}(\mathbb R,\mathbb R)$ be differentiable. Show that $f'$ has a zero in $(0,2\pi)$.

\vspace{0.5em}
\hrule
\vspace{0.8em}
\textbf{Output:} \\
\begin{lstlisting}[language=lean4]
import Mathlib

theorem ma_proofbench_l1_83 {f : ℝ → ℝ} (hper : Function.Periodic f (2 * Real.pi)) (hdiff : Differentiable ℝ f) :
    ∃ c ∈ Set.Ioo 0 (2 * Real.pi), deriv f c = 0 := by
  have πpos : 0 < Real.pi := Real.pi_pos
  have hlt : (0 : ℝ) < 2 * Real.pi := by linarith
  have hcont : ContinuousOn f (Set.Icc 0 (2 * Real.pi)) :=
    hdiff.continuous.continuousOn
  have hdiff_on : DifferentiableOn ℝ f (Set.Ioo 0 (2 * Real.pi)) :=
    hdiff.differentiableOn
  have h_eq : f (2 * Real.pi) = f 0 := by simpa using hper 0
  rcases exists_deriv_eq_slope f hlt hcont hdiff_on with ⟨c, hc, hc'⟩
  rw [h_eq, sub_self, zero_div] at hc'
  exact ⟨c, hc, hc'⟩
\end{lstlisting}

\vspace{0.5em}
\textbf{Analysis:} \\
The model successfully proves the existence of a zero for the derivative of a differentiable periodic function. It accurately identifies and applies the Mean Value Theorem (\texttt{exists\_deriv\_eq\_slope}) from Mathlib. By verifying the prerequisites of continuity and differentiability on the appropriate intervals, it cleverly uses the periodicity condition to deduce a zero slope, effectively concluding the proof.
\end{examplebox}

\begin{examplebox}[app:correct_answer_3]{Success Case (Gemini 3.1 Pro)}
\textbf{Question:} \\
Let $H$ be a Hilbert space and let $U$ be a bounded linear operator acting on $H$.
If $U$ is a unitary operator, and $\sigma(U)$ denotes the spectrum of $U$, prove that $\sigma(U)\subseteq\{z\in\mathbb{C}:|z|=1\}$.

\vspace{0.5em}
\hrule
\vspace{0.8em}
\textbf{Output:} \\
\begin{lstlisting}[language=lean4]
import Mathlib

theorem ma_proofbench_l2_73 (H : Type*) [NormedAddCommGroup H] [InnerProductSpace ℂ H] [CompleteSpace H]
    (U : H →L[ℂ] H)
    (h₁ : U.adjoint * U = 1) (h₂ : U * U.adjoint = 1) :
    spectrum ℂ U ⊆ {z : ℂ | ‖z‖ = 1} := by
  -- In the C*-algebra of continuous linear maps `H →L[ℂ] H`, `star U` is definitionally `U.adjoint`
  have h1_star : star U * U = 1 := h₁
  have h2_star : U * star U = 1 := h₂
  
  -- We construct the proof that U belongs to the unitary submonoid
  have hu : U ∈ unitary (H →L[ℂ] H) := ⟨h1_star, h2_star⟩
  let u : unitary (H →L[ℂ] H) := ⟨U, hu⟩
  
  -- Apply the general C*-algebra theorem that bounds the spectrum of a unitary element 
  -- to be within the metric sphere of radius 1 centered at 0.
  have h_spec : spectrum ℂ (u : H →L[ℂ] H) ⊆ Metric.sphere (0 : ℂ) 1 :=
    unitary.spectrum_subset_circle u
    
  -- Now we simply translate the `Metric.sphere` condition to the goal's norm condition
  intro z hz
  have hz' := h_spec hz
  
  -- `z ∈ Metric.sphere 0 1` means `dist z 0 = 1`. 
  -- Since `dist z 0 = ‖z‖`, it simplifies exactly to `‖z‖ = 1`.
  rw [Metric.mem_sphere, dist_zero_right] at hz'
  exact hz'
\end{lstlisting}

\vspace{0.5em}
\textbf{Analysis:} \\
This example demonstrates the model's proficiency in handling advanced functional analysis concepts. To bound the spectrum of a unitary operator on a Hilbert space, the model avoids reasoning from first principles and instead expertly navigates Mathlib's C*-algebra API. It explicitly constructs the unitary submonoid to apply the general theorem \texttt{unitary.spectrum\_subset\_circle}, and then smoothly translates the topological metric condition (\texttt{Metric.sphere}) into the algebraic norm condition required by the goal.
\end{examplebox}
\subsection{Failed Proofs}

\begin{examplebox}[app:failed_proof_1]{Failure Case: Mathlib Hallucination (DeepSeek-Prover-V2-671B)}
\textbf{Question:} \\
Verify that $g: \mathbb{R}^2 \setminus \{0\} \to \mathbb{R}$ defined by
\[
g(x) := \log|x|,
\]
is harmonic in $\mathbb{R}^2 \setminus \{0\}$.

\vspace{0.5em}
\hrule
\vspace{0.8em}
\textbf{Output:} \\
\begin{lstlisting}[language=lean4, escapeinside={(*@}{@*)}]
import Mathlib

theorem ma_proofbench_l1_50 : InnerProductSpace.HarmonicOnNhd (fun x : EuclideanSpace ℝ (Fin 2) => Real.log ‖x‖) ({0}ᶜ) := by
  have h_main : ∀ (x : EuclideanSpace ℝ (Fin 2)), x ∈ ({0}ᶜ : Set (EuclideanSpace ℝ (Fin 2))) → ∃ (u : Set (EuclideanSpace ℝ (Fin 2))) (hu : u ∈ (*@$\mathcal{N}$@*)[({0}ᶜ : Set (EuclideanSpace ℝ (Fin 2)))] x), (*@\colorbox{pink}{\textbf{InnerProductSpace.HarmonicOn}}@*) (fun x : EuclideanSpace ℝ (Fin 2) => Real.log ‖x‖) u := by
    intro x hx
    have hx' : x ≠ 0 := by simpa using hx
    use ({0}ᶜ : Set (EuclideanSpace ℝ (Fin 2)))
    use?
    (*@\colorbox{pink}{\textbf{sorry}}@*)
  intro x hx
  have h₁ : ∃ (u : Set (EuclideanSpace ℝ (Fin 2))) (hu : u ∈ (*@$\mathcal{N}$@*)[({0}ᶜ : Set (EuclideanSpace ℝ (Fin 2)))] x), (*@\colorbox{pink}{\textbf{InnerProductSpace.HarmonicOn}}@*) (fun x : EuclideanSpace ℝ (Fin 2) => Real.log ‖x‖) u := by
    apply h_main
    exact hx
  obtain ⟨u, hu, h₂⟩ := h₁
  exact ⟨u, hu, h₂⟩
\end{lstlisting}

\vspace{0.5em}
\textbf{Analysis:} \\
This proof fails due to a \textbf{Mathlib Hallucination}. The model fabricates a nonexistent theorem identifier, \texttt{InnerProductSpace.HarmonicOn}. In Mathlib 4, the correct identifier for this context should be \texttt{InnerProductSpace.HarmonicOnNhd}. Additionally, the model is unable to resolve the neighborhood filter logic and ultimately resorts to an explicit \texttt{sorry} to bypass the sub-goal.
\end{examplebox}

\begin{examplebox}[app:failed_proof_2]{Failure Case: Type System Error (DeepSeek-Prover-V2-7B)}
\textbf{Question:} \\
Suppose $\nu,\nu_{1},\nu_{2}$ are signed measures on $(X,\mathcal{M})$ and $\mu$ a (positive) measure on $\mathcal{M}$. Prove $\nu_{1}\perp \nu_{2}$ implies $|\nu_{1}|\perp |\nu_{2}|$.

\vspace{0.5em}
\hrule
\vspace{0.8em}
\textbf{Output:} \\
\begin{lstlisting}[language=lean4, escapeinside={(*@}{@*)}]
import Mathlib

theorem ma_proofbench_l1_98 {X : Type*} [MeasurableSpace X]
    (ν₁ ν₂ : MeasureTheory.SignedMeasure X)
    (h : MeasureTheory.VectorMeasure.MutuallySingular ν₁ ν₂) :
    (MeasureTheory.Measure.MutuallySingular ν₁.totalVariation ν₂.totalVariation) := by
  have h_main : MeasureTheory.Measure.MutuallySingular ν₁.totalVariation ν₂.totalVariation := by
    -- Use the fact that ν₁ and ν₂ are mutually singular to find a set B such that ν₁(B) = 0 and ν₂(Bᶜ) = 0
    have h₁ : ∃ (B : Set X), MeasurableSet B ∧ ν₁ B = 0 ∧ ν₂ Bᶜ = 0 := by
      exact h
    obtain ⟨B, hB_m, hB₁, hB₂⟩ := h₁
    -- Show that the total variation of ν₁ at B is 0
    have h₂ : ν₁.totalVariation B = 0 := by
      -- Compute the total variation of ν₁ at B
      have h₃ : (*@\colorbox{pink}{\textbf{$\nu_1$.totalVariation B = |$\nu_1$ B|}}@*) := by
        -- Use the definition of total variation for a signed measure
        simp [MeasureTheory.SignedMeasure.totalVariation, abs_eq_self.mpr (le_of_lt (show (0 : ℝ≥0) < 1 by norm_num))]
      rw [h₃]
      -- Simplify |ν₁ B| since ν₁ B = 0
      have h₄ : |ν₁ B| = 0 := by
        simp [hB₁]
        <;> aesop
      rw [h₄]
    -- Show that the total variation of ν₂ at Bᶜ is 0
    have h₃ : ν₂.totalVariation Bᶜ = 0 := by
      -- Compute the total variation of ν₂ at Bᶜ
      have h₄ : (*@\colorbox{pink}{\textbf{$\nu_2$.totalVariation B$^{\mathrm{c}}$ = |$\nu_2$ B$^{\mathrm{c}}$|}}@*) := by
        -- Use the definition of total variation for a signed measure
        simp [MeasureTheory.SignedMeasure.totalVariation, abs_eq_self.mpr (le_of_lt (show (0 : ℝ≥0) < 1 by norm_num))]
      rw [h₄]
      -- Simplify |ν₂ Bᶜ| since ν₂ Bᶜ = 0
      have h₅ : |ν₂ Bᶜ| = 0 := by
        simp [hB₂]
        <;> aesop
      rw [h₅]
    -- Conclude that ν₁.totalVariation and ν₂.totalVariation are mutually singular
    refine' ⟨B, hB_m, _⟩
    simp_all [MeasureTheory.Measure.MutuallySingular, h₂, h₃]
    <;> aesop
  exact h_main
\end{lstlisting}

\vspace{0.5em}
\textbf{Analysis:} \\
This represents a classic \textbf{Type System Error}. The model attempts to directly equate \texttt{$\nu_1$.totalVariation B} with \texttt{|$\nu_1$ B|}. However, the former evaluates to a measure value of type \texttt{ENNReal} (extended non-negative reals), whereas the latter evaluates to a standard metric value of type $\mathbb{R}$. Placing these two disparate types on opposite sides of an equality operator triggers a strict type mismatch failure during compiler verification.
\end{examplebox}

\begin{examplebox}[app:failed_proof_3]{Failure Case: Incomplete Proof (Claude Sonnet 4.6)}
\textbf{Question:} \\
The function $f\in C^{1}(\mathbb{R}^{2},\mathbb{R})$ satisfies $\partial_{1}f=\partial_{2}f$ and $f(0,0)=0$. Show that there is a $g\in C(\mathbb{R}^{2},\mathbb{R})$ such that $f(x,y)=g(x,y)(x+y)$ for $(x,y)\in\mathbb{R}^{2}$.

\vspace{0.5em}
\hrule
\vspace{0.8em}
\textbf{Output:} \\
\begin{lstlisting}[language=lean4, escapeinside={(*@}{@*)}]
import Mathlib

theorem ma_proofbench_l1_02 {f : ℝ × ℝ → ℝ} (hf : ContDiff ℝ 1 f)
    (hpar : ∀ p : ℝ × ℝ, (fderiv ℝ f p) (1, 0) = (fderiv ℝ f p) (0, 1))
    (h0 : f (0, 0) = 0) :
    ∃ g : ℝ × ℝ → ℝ, Continuous g ∧
      ∀ x y : ℝ, f (x, y) = g (x, y) * (x + y) := by
  let g : ℝ × ℝ → ℝ := fun p =>
    if p.1 + p.2 = 0 then
      (fderiv ℝ f (0, 0)) (1, 1) / 2
    else
      f p / (p.1 + p.2)
  use g
  constructor
  · -- Continuity of g
    have hf_cont : Continuous f := hf.continuous
    have hf_diff : Differentiable ℝ f := hf.differentiable le_rfl
    (*@\colorbox{pink}{\textbf{sorry}}@*)
  · intro x y
    by_cases h : x + y = 0
    · simp only [g, h, ite_true, mul_zero]
      (*@\colorbox{pink}{\textbf{sorry}}@*)
    · simp only [g, h, ite_false]
      field_simp
\end{lstlisting}

\vspace{0.5em}
\textbf{Analysis:} \\
This output exemplifies an \textbf{Incomplete Proof} characterized by explicit \texttt{sorry} insertion. While the model correctly defines the piecewise function $g(x,y)$ to satisfy the conditions, it struggles with the substantial logical leaps required to prove the continuity of this piecewise construction at the singularity boundary ($x+y=0$). Consequently, it abandons the proof by inserting \texttt{sorry} placeholders.
\end{examplebox}



                  


\begin{examplebox}[app:failed_proof_5]{Failure Case: Lean Syntax Errors (DeepSeek-V3.2-Thinking)}
\textbf{Question:} \\
Let $E$ be an n.v.s. and let $C\subset E$ be convex. Deduce that $\overline{C}=\overline{\operatorname{Int}C}$ whenever $\operatorname{Int}C\ne\varnothing.$

\vspace{0.5em}
\hrule
\vspace{0.8em}
\textbf{Output:} \\
\begin{lstlisting}[language=lean4, escapeinside={(*@}{@*)}]
import Mathlib
open Set

theorem ma_proofbench_l1_88 {E : Type*} [NormedAddCommGroup E] [NormedSpace ℝ E]
    {C : Set E} (hC : Convex ℝ C) (hInt : (interior C).Nonempty) :
    closure (interior C) = closure C := by
  apply Set.Subset.antisymm
  · exact closure_mono (interior_subset C)
  · intro x hx
    rcases hInt with ⟨y, hy⟩
    set f : ℝ × E → E := fun p => (1 - p.1) (*@$\cdot$@*) y + p.1 (*@$\cdot$@*) p.2 with hf_def
    have h_cont : Continuous f := by
      unfold f
      continuity
    have h_cont_at : ContinuousAt f (1, x) := h_cont.continuousAt
    have h_closure : (1, x) ∈ closure (Ioo (0 : ℝ) 1 ×ˢ C) := by
      rw [closure_prod_eq]
      constructor
      · have : (1 : ℝ) ∈ closure (Ioo (0 : ℝ) 1) := by
          rw [closure_Ioo (by norm_num)]
          exact ⟨by norm_num, by norm_num⟩
        exact this
      · exact hx
    have h_image : f '' (Ioo (0 : ℝ) 1 ×ˢ C) ⊆ interior C := by
      rintro _ ⟨⟨(*@\colorbox{pink}{\textbf{$\lambda$}}@*), w⟩, ⟨⟨hλ1, hλ2⟩, hw⟩, rfl⟩
      unfold f
      have H : (*@\colorbox{pink}{\textbf{$\lambda$}}@*) (*@$\cdot$@*) w + (1 - (*@\colorbox{pink}{\textbf{$\lambda$}}@*)) (*@$\cdot$@*) y ∈ interior C := 
        hC.combo_interior_self_subset_interior hw hy (by linarith) (by linarith) (by ring)
      simpa [add_comm] using H
    have H1 : f (1, x) ∈ closure (f '' (Ioo (0 : ℝ) 1 ×ˢ C)) :=
      h_cont_at.mem_closure_image h_closure
    have H2 : closure (f '' (Ioo (0 : ℝ) 1 ×ˢ C)) ⊆ closure (interior C) :=
      closure_mono h_image
    have : f (1, x) = x := by simp [f]
    rw [this] at H1
    exact H2 H1
\end{lstlisting}

\vspace{0.5em}
\textbf{Analysis:} \\
This proof suffers from an Illegal Identifier syntax error. In the destructuring step \texttt{rintro \_ $\langle\langle\lambda$, w$\rangle\rangle$, ...}, the model attempts to bind a real number to the variable name $\lambda$. However, $\lambda$ (\texttt{\char`\\lambda}) is a strictly reserved system keyword in Lean (utilized for lambda abstractions or anonymous functions). Utilizing it as a local variable identifier violates Lean 4 lexical rules and immediately breaks the parser.
\end{examplebox}
\section{Prompts} \label{appendix_prompts}

\begin{examplebox}[app:prompt_prover]{Prompt for Goedel-Prover-V2 and DeepSeek-Prover-V2}
\begin{Verbatim}[breaklines=true, breakanywhere=true]
Complete the following Lean 4 code:

```lean4
{formal_statement}
```
Before producing the Lean 4 code to formally prove the given theorem, provide a detailed proof plan outlining the main proof steps and strategies.
The plan should highlight key ideas, intermediate lemmas, and proof structures that will guide the construction of the final formal proof.
\end{Verbatim}
\end{examplebox}

\begin{examplebox}[app:kimina_prover]{Prompt for Kimina-Prover}
\begin{Verbatim}[breaklines=true, breakanywhere=true]
You are an expert in mathematics and proving theorems in Lean 4. Think about and solve the following problem step by step in Lean 4.
Problem:{informal_statement}

Formal statement:

```lean4
{formal_statement}
```
\end{Verbatim}
\end{examplebox}

\begin{examplebox}[app:general_models]{Prompt for General-Purpose Models}
\begin{Verbatim}[breaklines=true, breakanywhere=true]
You are an expert in Lean 4 and Mathematics. Please finish the following proof in Lean4 code.
Do not change the original statement. Copy the final statement to prove exactly.
Please include the complete header (including imports and namespaces) so that your code can pass the Lean4 compiler. Please solve the statement step by step and provide your complete Lean4 code between ```lean4 and ``` after careful reasoning.
The statement for you to complete is:

```lean4
{formal_statement}
```
\end{Verbatim}
\end{examplebox}

\begin{examplebox}[app:prompt_informal_judge]{Prompt for GPT-5.5 Informal Proof Grading}
\begin{Verbatim}[breaklines=true, breakanywhere=true]
## Instruction

Your task is to evaluate the quality of a solution to a problem. The problem may ask for a proof of statement, or ask for an answer. If finding an answer is required, the solution should present the answer, and it should also be a rigorous proof of that answer being valid.

Please evaluate the solution and score it according to the following criteria:
- If the solution is completely correct, with all steps executed properly and clearly demonstrated, then the score is 1
- If the solution is generally correct, but with some details omitted or minor errors, then the score is 0.5
- If the solution does not actually address the required problem, contains fatal errors, or has severe omissions, then the score is 0
- Additionally, referencing anything from any paper does not save the need to prove the reference. It's okay IF AND ONLY IF the solution also presents a valid proof of the reference argument(s); otherwise, if the solution omits the proof or if the proof provided is not completely correct, the solution should be scored according to the criteria above, and definitely not with a score of 1

Please carefully reason out and analyze the quality of the solution below, and in your final response present a detailed evaluation of the solution's quality followed by your score. Therefore, your response should be in the following format:

Here is my evaluation of the solution:
... // Your evaluation here. You are required to present in detail the key steps of the solution or the steps for which you had doubts regarding their correctness, and explicitly analyze whether each step is accurate: for correct steps, explain why you initially doubted their correctness and why they are indeed correct; for erroneous steps, explain the reason for the error and the impact of that error on the solution.

Based on my evaluation, the final overall score should be:
\boxed{...} // where ... should be the final overall score (0, 0.5, or 1, and nothing else) based on the above criteria

---

Here is your task input:

## Problem
{statement}

## Solution
{proof}
\end{Verbatim}
\end{examplebox}

\end{document}